\documentclass[sigconf]{acmart}

\usepackage{float}
\usepackage{booktabs}
\usepackage{amsmath}
\usepackage{amsthm}
\usepackage{amssymb}
\usepackage{graphicx}
\usepackage{balance} 
\usepackage{needspace}
\usepackage{booktabs}

\floatstyle{ruled}
\newfloat{algorithm}{tbp}{loa}
\providecommand{\algorithmname}{Algorithm}
\floatname{algorithm}{\protect\algorithmname}

\usepackage{times}
\usepackage{color}
\definecolor{gray}{gray}{0.96} 

\usepackage[caption=false]{subfig}

\usepackage{listings}

\renewcommand\footnotetextcopyrightpermission[1]{} 
\makeatletter
\def\runningfoot{\def\@runningfoot{}}
\def\firstfoot{\def\@firstfoot{}}
\makeatother 

\setcopyright{none}

\begin{document}

%


\title{TEASER: Early and Accurate Time Series Classification}

\author{Patrick Sch\"afer}
\affiliation{%
	\institution{Humboldt University of Berlin, Germany}
}
\email{patrick.schaefer@hu-berlin.de}

\author{Ulf Leser}
\affiliation{
	\institution{Humboldt University of Berlin, Germany}
}
\email{leser@informatik.hu-berlin.de}

\renewcommand{\shortauthors}{P. Sch\"afer and U. Leser}

\begin{abstract} 
Early time series classification (eTSC) is the problem of classifying a time series after as few measurements as possible with the highest possible accuracy. The most critical issue of any eTSC method is to decide when enough data of a time series has been seen to take a decision: Waiting for more data points usually makes the classification problem easier but delays the time in which a classification is made; in contrast, earlier classification has to cope with less input data, often leading to inferior accuracy. The state-of-the-art eTSC methods compute a fixed optimal decision time assuming that every times series has the same defined start time (like turning on a machine). However, in many real-life applications measurements start at arbitrary times (like measuring heartbeats of a patient), implying that the best time for taking a decision varies heavily between time series. We present TEASER, a novel algorithm that models eTSC as a two two-tier classification problem: In the first tier, a classifier periodically assesses the incoming time series to compute class probabilities. However, these class probabilities are only used as output label if a second-tier classifier decides that the predicted label is reliable enough, which can happen after a different number of measurements. In an evaluation using 45 benchmark datasets, TEASER is two to three times earlier at predictions than its competitors while reaching the same or an even higher classification accuracy. We further show TEASER's superior performance using real-life use cases, namely energy monitoring, and gait detection.
\end{abstract}


\maketitle

\sloppy

\section{Introduction}

A time series (TS) is a collection of values sequentially ordered in time. One strong force behind their rising importance is the increasing use of sensors for automatic and high resolution monitoring in domains like smart homes~\cite{jerzak2014debs}, starlight observations~\cite{protopapas2006finding}, machine surveillance~\cite{mutschler2013debs}, or smart grids~\cite{hobbs1999analysis, WindPower}. Time series classification (TSC) is the problem of assigning one of a predefined class to a time series, like recognizing the electronic device producing a certain temporal pattern of energy consumption~\cite{gao2014plaid,gisler2013appliance} or classifying a signal of earth motions as either an earthquake or a bypassing lorry~\cite{perol2018convolutional}. 
\begin{figure}[t]
	\includegraphics[width=1\columnwidth]{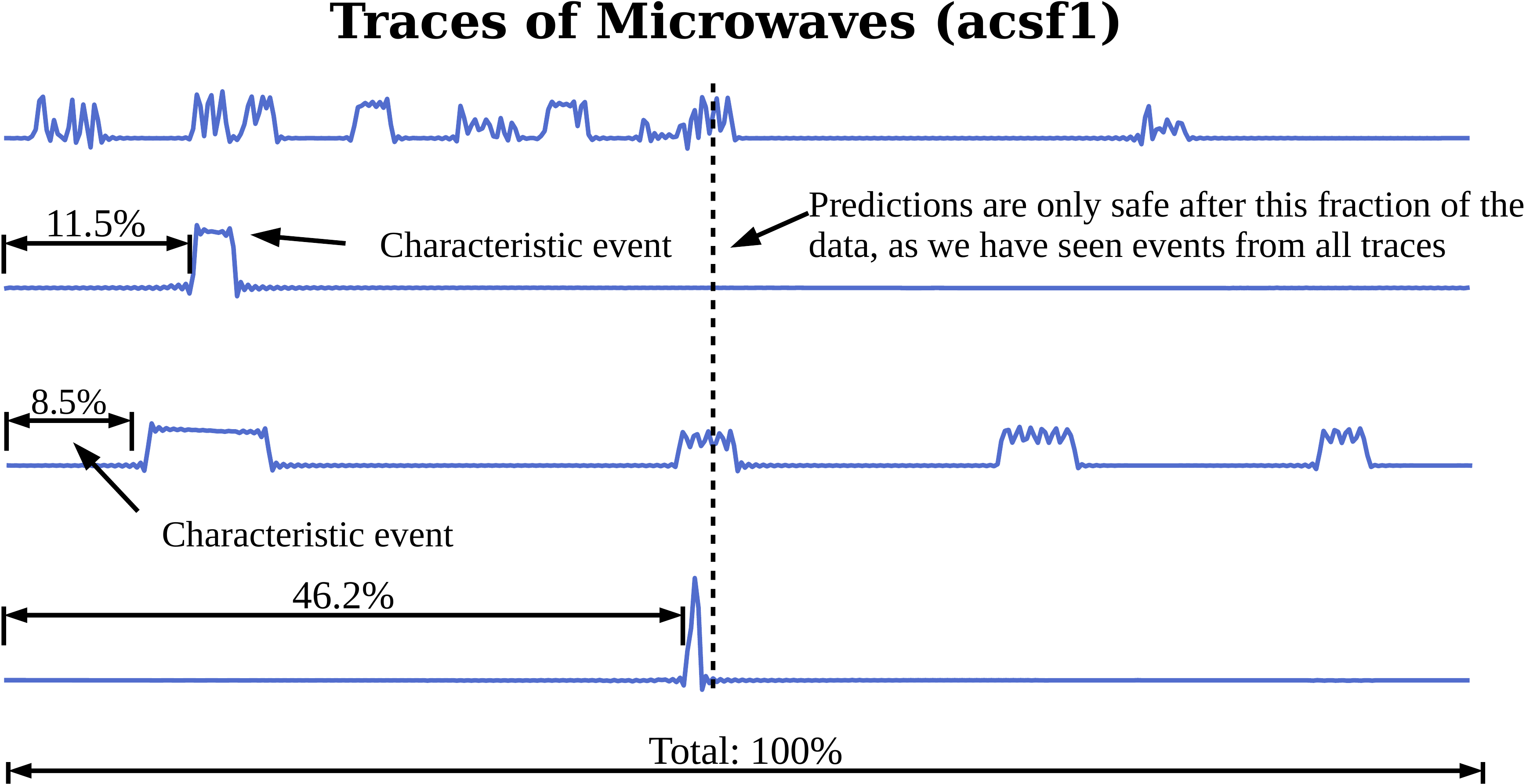}
	
	\caption{Traces of microwaves taken from~\protect\cite{gisler2013appliance}. The operational state of the microwave starts between $5\%$ and $50\%$ of the whole trace length. To have at least one event (typically a high burst of energy consumption) for each microwave, the threshold has to be set to that of the latest seen operational state (after seeing more than $46.2\%$).\label{fig:Traces-of-a}
	}
\end{figure}
Conventional TSC works on time series of a given, fixed length and assumes access to the entire input at classification time. In contrast, \emph{early time series classification (eTSC)}, which we study in this work, tries to solve the TSC problem after seeing as few measurements as possible~\cite{xing2012early}. This need arises when the classification decision is time-critical, for instance to prevent damage (the earlier a warning system can predict an earthquake from seismic data~\cite{perol2018convolutional}, the more time there is for preparation), to speed-up diagnosis (the earlier an abnormal heart-beat is detected, the more time there is for prevention of fatal attacks~\cite{griffin2001toward}), or to protect markets and systems (the earlier a crisis of a particular stock is detected, the faster it can be banned from trading~\cite{ghalwash2014utilizing}).

\emph{eTSC} has two goals: Classifying TS early and with high accuracy. However, these two goals are contradictory in nature: The earlier a TS has to be classified, the less data is available for this classification, usually leading to lower accuracy. In contrast, the higher the desired classification accuracy, the more data points have to be inspected and the later eTSC will be able to take a decision. Thus, a critical issue in any eTSC method is the determination of the point in time at which an incoming TS can be classified. 
State-of-the-art methods in eTSC~\cite{xing2012early,xing2011extracting, mori2017reliable} assume that all time series being classified have a defined start time. Consequently, these methods assume that characteristic patterns appear roughly at the same offset in all TS, and try to learn the fixed fraction of the TS that is needed to make high accuracy predictions, i.e., when the accuracy of classification most likely is close to the accuracy on the full TS. However, in many real life applications this assumption is wrong. For instance, sensors often start their observations at arbitrary points in time of an essentially indefinite time series. Intuitively, existing methods expect to see a TS from the point in time when the observed system starts working, while in many applications, this system has already been working for an unknown period of time when measurements start. In such settings, it is suboptimal to wait for a fixed number of measurements; instead, the algorithm should wait for the characteristic patterns, which may occur early (in which case an early classification is possible) or later (in which case the eTSC algorithm has to wait longer). As an example, Figure~\ref{fig:Traces-of-a} illustrates traces for the operational state of microwaves~\cite{gisler2013appliance}. Observations started while the microwaves were already under power; the concrete operational state, characterized by high bursts of energy consumption, happened after $5\%$ to $50\%$ of the whole trace (amounting to 1 hour). Current eTSC methods trained on this data would always wait for 30mins, because this was the last time they had seen the important event in the training data. But actually most TS could be classified safely much earlier; instead of assuming a fixed classification time, an algorithm should adapt its decision time to the individual time series. 

In this paper we present TEASER, a \emph{Two-tier Early and Accurate Series classifiER}, that is robust regarding the start time of a TS's recording. It models eTSC as a two-tier classification problem (see Figure~\ref{fig:Early-Accurate-Pipeline}). In the first tier, a slave classifier periodically assesses the input series and computes class probabilities. In the second tier, a master classifier takes the series of class probabilities of the slave as input and computes a binary decision on whether to report these as final result or continue with the observation. As such, TEASER does not presume a fixed starting time of the recordings nor does it rely on a fixed decision time for predictions, but takes its decisions whenever it confident of its prediction. 
On a popular benchmark of $45$ datasets~\cite{UCRClassification}, TEASER is two to three times as early while keeping a competitive, and for some datasets reaching an even higher level of accuracy, when compared to the state of the art~\cite{mori2017early,mori2017reliable, xing2011extracting, parrish2013classifying, xing2012early}. Overall, TEASER achieves the highest average accuracy, lowest average rank, and highest number of wins among all competitors. We furthermore evaluate TEASER's performance on the basis of real use-cases, namely device-classification of energy load monitoring traces, and classification of walking motions into normal and abnormal.

The rest of the paper is organized as follows: In Section 2 we formally describe the background of eTSC. Section 3 introduces TEASER and its building blocks. Section 4 presents evaluation results including benchmark data and real use cases. Section 5 discusses related work and Section 6 presents the conclusion.

\section{Background: Time Series and eTSC}\label{sec:definitions}

\begin{table}[t]
	\begin{centering}
		\begin{tabular}{cl}
			\toprule 
			Symbol & Meaning\tabularnewline
			\midrule
			\midrule 
			$sc_{i}$ / $mc_{i}$ & a slave / master classifier at the $i$'th snapshot\tabularnewline
			\midrule 
			$S$ & the number master/slave classifiers\tabularnewline
			\midrule 
			$w$ & a user-defined interval length\tabularnewline
			\midrule 
			$s_i$ & the snapshot length, with $s_i=i\cdot w$\tabularnewline
			\midrule 		
			$n$ & the time series length\tabularnewline
			\midrule 
			$N$ & the number of samples\tabularnewline
			\midrule 
			k & number of classes\tabularnewline
			\midrule 
			$p(c_{i})$ & class probability by the slave classifier\tabularnewline
			\midrule 
			Y & all class labels, $Y=\left\{ c_{1},\ldots c_{k}\right\} $\tabularnewline
			\midrule 
			$c_{i}$ & i'th class label in $Y$\tabularnewline
			\bottomrule
		\end{tabular}
		\par\end{centering}
	\caption{Symbols and Notations.\label{tab:Symbols-and-Notations.}}
\end{table}

In this section, we formally introduce time series (TS) and early time series classification (eTSC). We also describe the typical learning framework used in eTSC.
\begin{definition}
A \emph{time series} $T$ is a sequence of $n \in \mathbb{N}$ real values, $T=(t_{1},\ldots,t_{n}),t_{i} \in \mathbb{R}$. The values are also called data points. A dataset $D$ is a collection of time series. 
\end{definition}
We assume that all TS of a dataset have the same sampling frequency, i.e., every i'th data point was measured at the same temporal distance from the first point. In accordance to all previous approaches~\cite{xing2012early,xing2011extracting, mori2017reliable}, we will measure earliness in the number of data points and from now on disregard the actual time of data points.
A central assumption of eTSC is that TS data arrives incrementally. If a classifier is to classify a TS after $s$ data points, it has access to these $s$ data points only. This is called a \emph{snapshot}.

\begin{definition}
	A \emph{snapshot} $T(s)=(t_{1},\ldots,t_{s})$ of a time series $T$, $k \leq n$, is the prefix of $T$ available for classification after seeing $s$ data points.
\end{definition}

In principle, an eTSC system could try to classify a time series after every new data point that was measured. However, it is more practical and efficient to call the eTSC only after the arrival of a fixed number of new data points~\cite{xing2012early,xing2011extracting, mori2017reliable}. We call this number the \emph{interval length} $w$. Typical values are $5, 10, 20,\dots$

eTSC is commonly approached as a supervised learning problem~\cite{mori2017early,mori2017reliable, xing2011extracting, parrish2013classifying, xing2012early}. Thus, we assume the existence of a set $D_{train}$ of training TS, where each one is assigned to one of a predefined set of class labels $Y=\left\{ c_{1},\ldots ,c_{k}\right\}$. The eTSC system learns a model from $D_{train}$ that can separate the different classes. Its performance is estimated by applying this model to all instances of a test set $D_{test}$. 

The quality of an eTSC system can be measured by different indicators. The \emph{accuracy} of an eTSC is calculated as the percentage of correct predictions of the test instances, where higher is better: 
\[
\textit{accuracy}=\frac{\textit{number of correct predictions}}{\left|D_{test}\right|}
\]
The earliness of an eTSC is defined as the mean number of data points $s$ after which a label is assigned, where lower is better: 
\[
\textit{earliness}=\frac{\sum_{T\epsilon D_{test}} \frac{s}{len(T)} }{\left|D_{test}\right|}
\]

We can now formally define the problem of eTSC.

\begin{figure}[t]
\includegraphics[width=1\columnwidth]{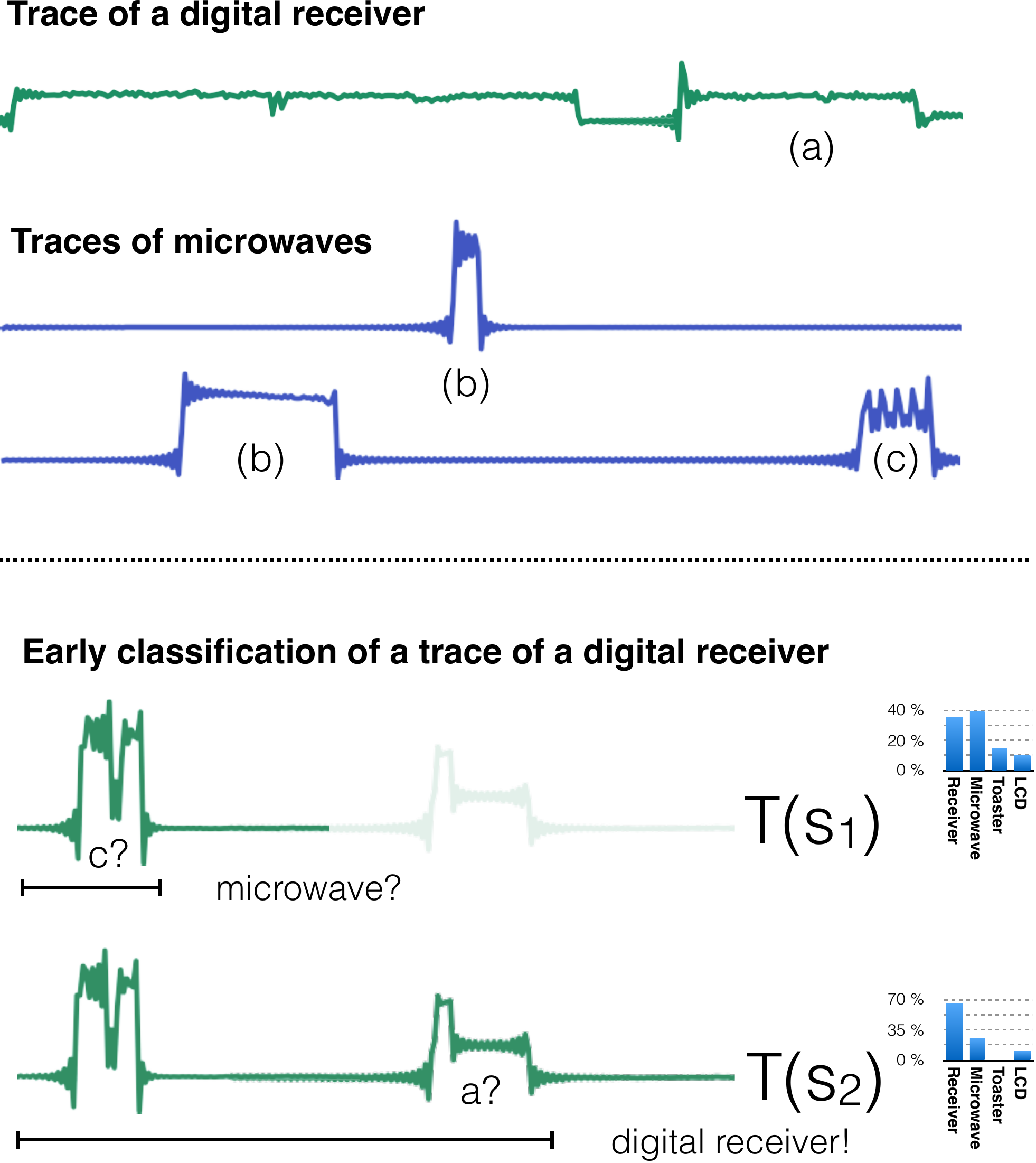}

\caption{eTSC on a trace of a digital receiver. The figure shows one traces of a digital receiver, and of microwaves on the top. These have three characteristic patterns (a) to (c). In the bottom part, eTSC is performed on a snapshot of the time series of a digital receiver. In its first snapshot it is easily confused with pattern (c) of a microwave. However, the trace later contains pattern (a) which is characteristic for a receiver.
\label{fig:ETSC}}
\end{figure}

\begin{definition}
	\emph{Early time series classification} (eTSC) is the problem of assigning all time series $T \in D_{test}$ a label from $Y$ as early and as accurate as possible. 
\end{definition}

eTSC thus has two optimization goals that are contradictory in nature, as later classification typically allows for more accurate predictions and vice versa. Accordingly, eTSC methods can be evaluated in different ways, such as comparing accuracies at a fixed-length snapshot (keeping earliness constant), comparing earliness at which a fixed accuracy is reached (keeping accuracy constant), or by combining these two measures. A popular choice for the latter is the harmonic mean of earliness and accuracy:

\[
HM=\frac{2\cdot(1-earliness)\cdot accuracy}{(1-earliness)+accuracy}
\]
An $HM$ of $1$ is equal to an earliness of $0\%$ and an accuracy of $100\%$. 
Figure~\ref{fig:ETSC} illustrates the problem of eTSC on a load monitoring task differentiating a \emph{digital receiver} from a \emph{microwave}. All traces have an underlying oscillating pattern and in total there are three important patterns (a), (b), (c) which are different among the appliances. The characteristic part of a \emph{receiver} trace is an energy burst with two plateaus (a), which can appear at different offsets. If an eTSC classifies a trace too early (Figure~\ref{fig:ETSC} second from bottom), the signal is easily confused with that of microwaves based on the similarity to the (c) pattern. 
However, if an eTSC always waits until the offset at which all training traces of \emph{microwaves} can be correctly classified, the first \emph{receiver} trace will be classified much later than possible (eventually after seeing the full trace). To achieve optimal earliness at high accuracy, an eTSC system must determine its decision times individually for each TS it analyses.

\section{Early \& Accurate TS Classification: TEASER}

\begin{figure*}[th]
	\includegraphics[width=2\columnwidth]{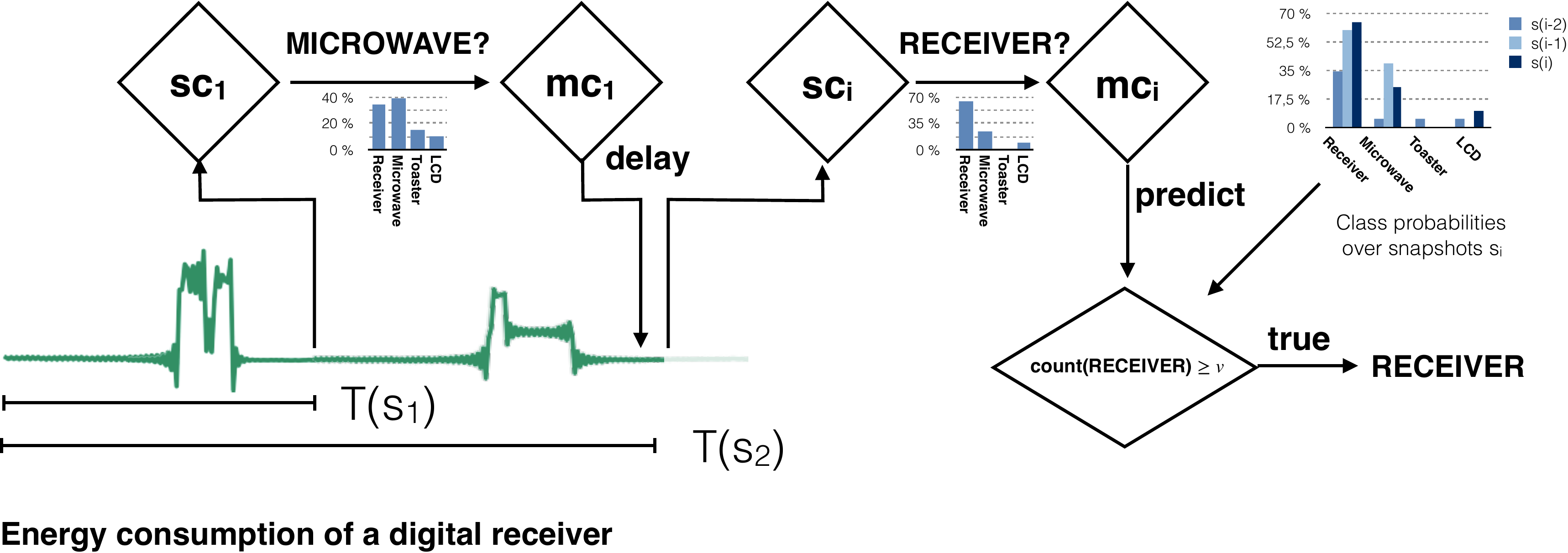}	
	\caption{TEASER is given a snapshot of an energy consumption time series. After seeing the first $s$ measurements, the first slave classifier $sc_1$ performs a prediction which the master classifier $ mc_1$ rejects due to low class probabilities. After observing the $i$'th interval which includes a characteristic energy burst, the slave classifier $sc_i$ (correctly) predicts RECEIVER, and the master classifier $mc_i$ eventually accepts this prediction. When the prediction of RECEIVER has been consistently derived $v$ times, it is output as final prediction.\label{fig:Early-Accurate-Pipeline}}
\end{figure*}

TEASER addresses the problem of finding optimal and individual decision times by following a two-tier approach. Intuitively, it trains a pair of classifiers for each snapshot~$s$: A \emph{slave classifier} computes class probabilities which are passed on to a \emph{master classifier} that decides whether these probabilities are high enough that a safe classification can be emitted. TEASER monitors these predictions and predicts a class $c$ if it occurred $v$ times in a row; the minimum length $v$ of a series of predictions is an important parameter of TEASER. Intuitively, the slave classifiers give their best prediction based on the data they have seen, whereas the master classifiers decide if these results can be trusted, and the final  filter suppresses spurious predictions.

Formally, let $w$ be the user-defined \emph{interval length} and let $n_{max}$ be the length of the longest time series in the training set $D_{train}$. We then extract snapshots  $T(s_i)=T[1 .. i \cdot w]$, i.e., time series snapshots of lengths $s_i = i \cdot w$. A TEASER model consists of a set of $S=[n_{max}/w]$ pairs of slave/master classifiers, trained on the snapshots of the TS in $D_{train}$ (see below for details). When confronted with a new time series, TEASER waits for the next $w$ data points to arrive and then calls the appropriate slave classifier which outputs probabilities for all classes. Next, TEASER passes these probabilities to the slave's paired master classifier which either returns a class label or \emph{NIL}, meaning that no decision could be derived. If the answer is a class label $c$ and this answer was also given for the last $v-1$ snapshots, TEASER returns $c$ as result; otherwise, it keeps waiting.

Before going into the details of TEASER's components, consider the example shown in Figure~\ref{fig:Early-Accurate-Pipeline}. The first slave classifier $sc_1$ falsely labels this trace of a \emph{digital receiver} as a \emph{microwave} (by computing a higher probability of the latter class than for the former class) after seeing the first $w$ data points. However, the master classifier $mc_{1}$ decides that this prediction is unsafe and TEASER continues to wait. After $i-1$ further intervals, the $i$'th pair of slave and master classifiers $sc_i$ and $mc_i$ are called. Because the TS contained characteristic patterns in the $i$'th interval, the slave now computes a high probability for the \emph{digital receiver} class, and the master decides that this prediction is safe. TEASER counts the number of consecutive predictions for this class and, if a threshold is passed, outputs the predicted class.  

Clearly, the interval length $w$ and the threshold $v$ are two important yet opposing parameters of TEASER. A smaller $w$ results in more frequent predictions, due to smaller prediction intervals. However, a classification decision usually only changes after seeing a sufficient number of novel data points; thus, a too small value for $w$ leads to series of very similar class probabilities at the slave classifiers, which may trick the master classifier. This can be compensated by increasing $v$. In contrast, a large value for $w$ leads to fewer predictions, where each one has seen more new data and thus is probably more reliable. For such settings, $v$ may be reduced without harming earliness or accuracy. 
In our experiments, we shall analyze the influence of $w$ on accuracy and earliness in Section~\ref{subsec:varying-interval-size}. In all experiments $v$ is treated as a hyper-parameter that is learned by performing a grid-search and maximizing \emph{HM} on the training dataset.

\subsection{Slave Classifier}\label{subsec:Slave-Classifier}

\begin{figure}[t]
	\includegraphics[width=1\columnwidth]{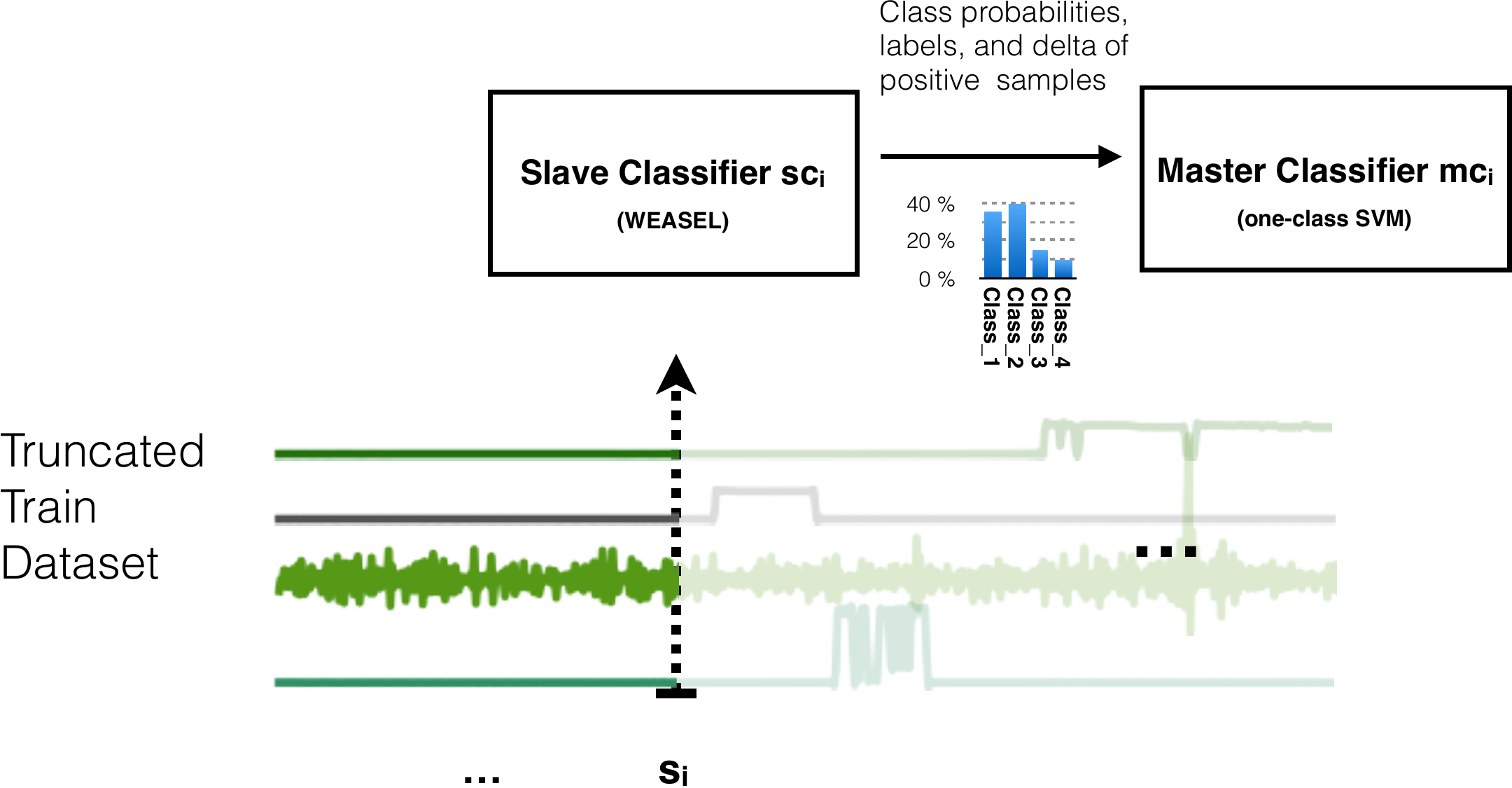}
	\caption{TEASER trains pairs of slave and master classifiers.
		The $i$'th slave classifier is trained on the time series truncated after
		time stamp $s_i$. The master classifier is trained on the
		class probabilities, and delta of the correctly predicted time series. \label{fig:Training-the-}}
\end{figure}

Each slave classifier $sc_i, \textit{ with } i \leq S$ is a full-fledged time series classifier of its own, trained to predict classes after seeing a fixed snapshot length. Given a snapshot $T(s_i)$ of length $s_i=i \cdot w$, the slave classifier $sc_i$ computes class probabilities 
$P(s_i) = \Bigl[ p\bigl(c_1(s_i)\bigr), \ldots , p\bigl(c_k(s_i)\bigr) \Bigr]$ 
for this time series for each of the predefined classes and determines the class $c(s_i)$ with highest probability. Furthermore, it computes the difference $\triangle d_i$ between the highest
\[
m_{i1}=\underset{j\epsilon[1\ldots k]}{arg\,max}\bigl\{p(c_j(s_i))\bigr\}
\]
and second highest 
\[
m_{i2}=\underset{j\epsilon[1\ldots k],j\neq m_1}{arg\,max}\bigl\{p(c_j(s_i))\bigr\}
\]
class probabilities: 
\[
\triangle d_i=p(c_{m_{i1}})-p(c_{m_{i2}})
\]
 
In TEASER, the most probable class label $c(s_i)$, the vector of class probabilities $P(s_i)$, and the difference $\triangle d(s_i)$ are passed as features to the paired $i$'th master classifier $mc_i$, which then has to decide if the prediction is reliable (see Figure~\ref{fig:Training-the-}) or not.

\subsection{Master Classifier}\label{subsec:Training-the-Set}

\begin{figure}[t]
	\includegraphics[width=1\columnwidth]{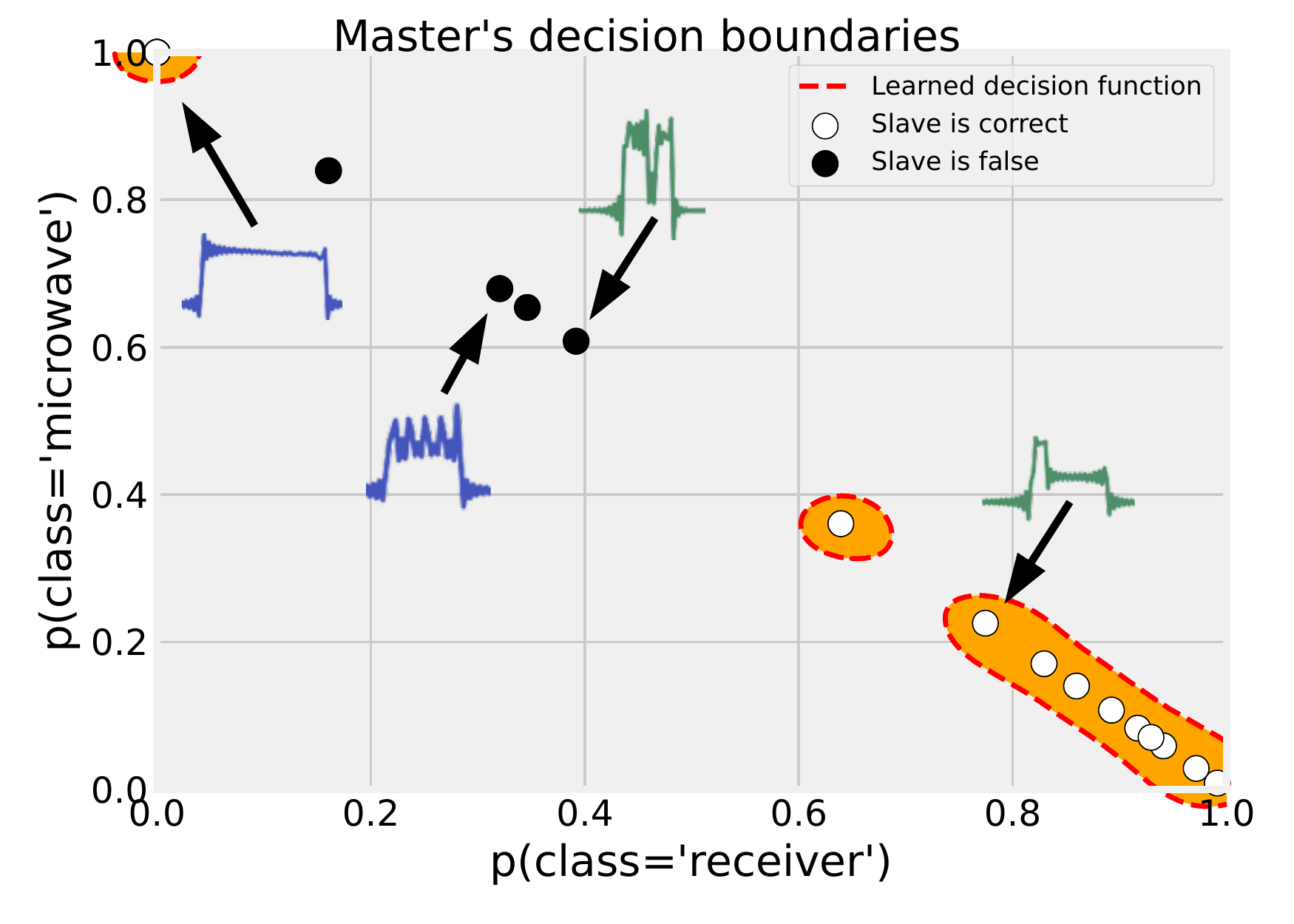}
	\caption{The master computes a hyper-sphere around the correctly predicted samples. A novel sample is accepted/rejected if it's probabilities fall into/outside the orange hypersphere.\label{fig:The-concept-behind}}
\end{figure}

\begin{figure}[t]
	\includegraphics[width=1\columnwidth]{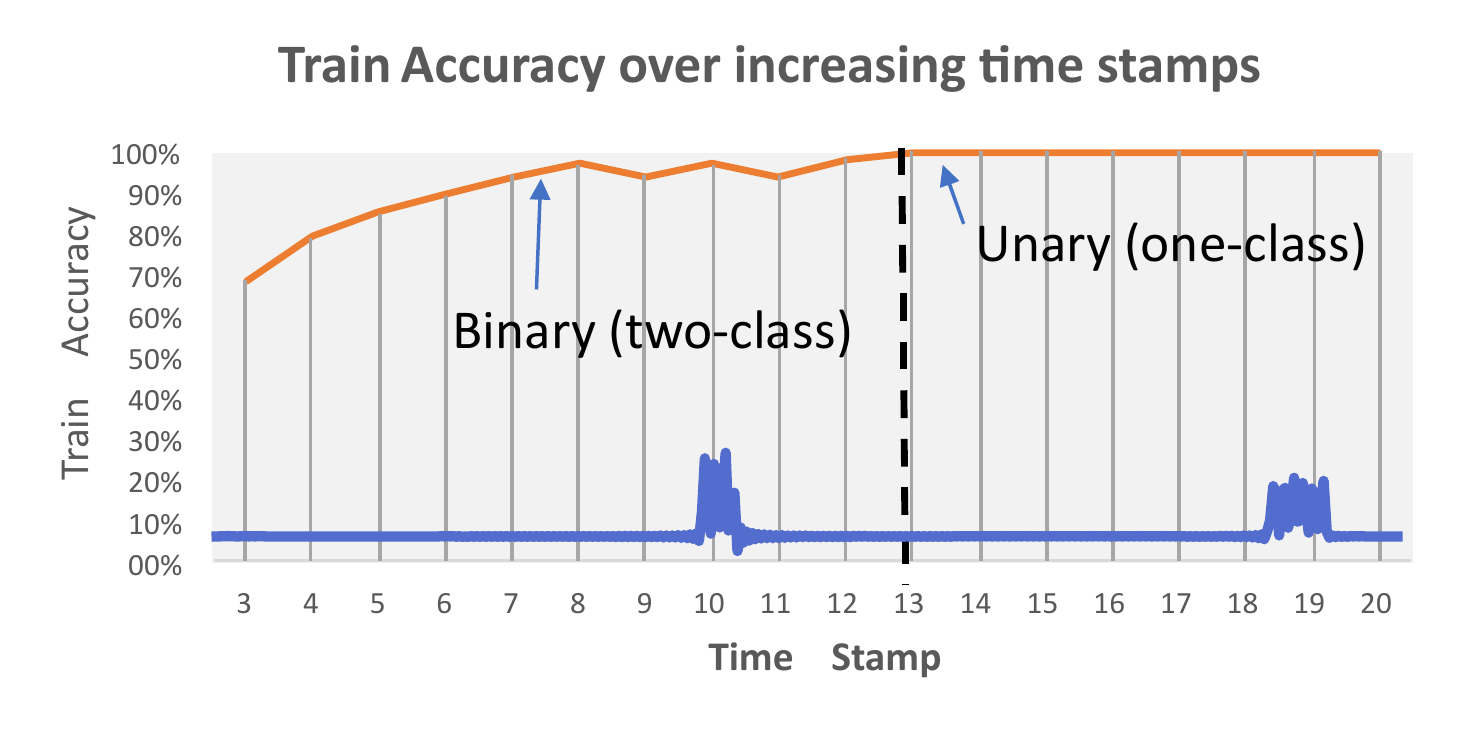}
	
	\caption{The accuracy of the slave classifier reaches 100\% after seeing $13$
		time stamps on the train data, resulting in one-class classification.\label{fig:Training-a-bimodal}}
\end{figure}

A master classifier $mc_i, \textit{ with } i \leq S$ in TEASER learns whether the results of its paired slave classifier should be trusted or not. We model this task as a classification problem in its own, where the $i$'th master classifier uses the results of the $i$'th slave classifier as features for learning its model (see Section~\ref{subsec:Training-Slave-Master} for the details on training). However, training this classifier is tricky. To learn accurate decisions, it needs to be trained with a sufficient number of correct and false predictions. However, the more accurate a slave classifier is, the less mis-classifications are produced, and the worse gets the expected performance of the paired master classifier. Figure~\ref{fig:Training-a-bimodal} illustrates this problem by showing a typical slave's train accuracy with an increasing number of data points. In this example, the slave classifiers start with an accuracy of around $70\%$ and reach $100\%$ quickly on the train data. Once the train accuracy reaches $100\%$, there are no negative samples left for the master classifier to train its decision boundary.

To overcome this issue, we use a so-called one-class classifier as master classifier~\cite{khan2009survey}. One-class classification refers to the problem of classifying positive samples in the absence of negative samples. It is closely related to, but non-identical to outlier/anomaly detection~\cite{cuturi2011autoregressive}. In TEASER, we use a one-class Support Vector Machine (oc-SVM)~\cite{scholkopf2001estimating} which does not determine a separating hyperplane between positive and negative samples but instead computes a hyper-sphere around the positively labeled samples with minimal dilation that has maximal distance to the origin. At classification time, all samples that fall outside this hypersphere are considered as negative samples; in our case, this implies that the master learns a model of positive samples and regards all results not fitting this model as negative. The major challenge is to determine a hyper-sphere that is neither too large nor too small to avoid false positives, leading to lower accuracy, or dismissals, which lead to delayed predictions. 

In Figure~\ref{fig:The-concept-behind} a trace is either labeled as \emph{microwave} or \emph{receiver}, and the master classifier learns that its paired slave is very precise at predicting \emph{receiver} traces but produces many false predictions for \emph{microwave} traces. Thus, only \emph{receiver} predictions with class probability above $p('receiver') \geq 0.6$, and microwaves above $p('microwave') \geq 0.95$ are accepted. As can be seen in this example, using a one-class SVM leads to very flexible decision boundaries.

\subsection{Training Slave and Master Classifiers \label{subsec:Training-Slave-Master}}

Consider a labeled set of time series $D_{train}$ with class labels $Y=\left\{ c_{1},\ldots, c_{k}\right\}$ and an interval length $w$. As before, $n_{max}$ is the length of the longest training instance. Then, the $i$'th pair of slave / master classifier is trained as follows:

\begin{enumerate}
	\item First, we truncate the train dataset $D_{train}$ to the prefix length determined by $i$ (snapshot $s_i$): 
	\[
	D_{train}(s_i)=\{T(s_i) ~|~ T \in D_{train}\}
	\]
	In Figure~\ref{fig:Training-the-} (bottom) the four TS are truncated.
	
	\item Next, these truncated snapshots are z-normalized. This is a critical step for training to remove any bias resulting from values that will only be available in the future. I.e., if a time series is first z-normalized like all UCR time series, and then a truncated snapshot is generated, this snapshot may not make use of the absolute values resulting from the z-normalization of the whole series (as opposed to \cite{mori2017early}). 
	
	\item The hyper-parameters of the slave classifier are trained on $D_{train}(s_i)$ using $10$-fold-cross validation. Using the derived hyper-parameters we can build the final slave classifier $sc_i$ producing its 3-tuple output $(c(s_i), P(s_i), \triangle d(s_i))$ for each $T \in D_{train}$ (Figure~\ref{fig:Training-the-} centre).
	
	\item To train the paired master classifier, we first remove all instances which were incorrectly classified by the slave. Assume that there were $N'\leq N$ correct predictions. We then train a one-class SVM on the $N'$training samples, where each sample is represented by the 3-tuple $(c(s_i), P(s_i), d(s_i)$ produced by the slave classifier. 
	
	\item Finally, we perform a grid-search over values $v\in\{1...5\}$ to find the threshold $v$ which yields the highest harmonic mean $HM$ of earliness and accuracy on $D_{train}$. 
	
\end{enumerate}

\begin{algorithm}[th]
	{\tiny}
	\begin{lstlisting}[basicstyle={\small\ttfamily},showstringspaces=false,tabsize=2]
	Input Trainig set: X_data
	Input Labels:	   Y_labels
	Input Time Stamps: {1,2,...,S}
	Returns: Slaves, Masters, v
	(1)	initialize array of slaves
	(2)	initialize array of masters
	(3)	for t in {1,2,...,S} do
	// z-normalized snapshorts
	(4)		X_snapshot_normed
	= z-norm(truncateAfter(X_data, t));
	(5)		c_probs, c_labels 
	= slaves[t].fit(X_snapshot_normed);
	// we keep the positive samples
	(6)		c_pos_probs = filter_correct(
	c_labels, train_labels, c_probs);
	// train a one-class classifier
	(7)		masters[t].fit(c_pos_probs);  
	// Maximize HM to find the best v
	(8)		v = grid_search(X_data, Y_labels, 
	slaves[t], masters[t]); 
	(9)	end for	
	(10)return (slaves, masters, v);
	\end{lstlisting}
	{\tiny}
	\caption{Training phase of TEASER using S time stamps, and a labeled train
		dataset. \label{alg:Training-phase-of}}
\end{algorithm}

In accordance to prior works~\cite{mori2017early, xing2011extracting, parrish2013classifying, xing2012early}, we consider the interval length $w$ to be a user-specified parameter. However, we will also investigate the impact of varying $w$ in Section \ref{subsec:varying-interval-size}. 

The pseudo-code for training TEASER is given in Algorithm~\ref{alg:Training-phase-of}. The aim of the training is to obtain $S$ pairs of slave/master classifiers, and the threshold $v$ for consecutive predictions. First, for all z-normalized snapshots (line~4), the slaves are trained and the predicted labels and class probabilities are kept (line~5). Prior to training the  master, incorrectly classified instances are removed (line~6). The feature vectors of correctly labeled samples are passed on to train the master (one-class SVM) classifier (line~7). Finally, an optimal value for $v$ is determined using grid-search.

\section{Experimental Evaluation}

We first evaluate TEASER using the $45$ datasets from the UCR archive that also have been used in prior works on eTSC~\cite{xing2012early, parrish2013classifying, xing2011extracting, mori2017reliable}. Each UCR dataset provides a train and test split set which we use unchanged. Note that most of these datasets were preprocessed to create approximately aligned patterns of equal length and scale~\cite{schafer2014towards}. Such an alignment is advantageous for methods that make use of a fixed decision time but also requires additional effort and introduces new parameters that must be determined, steps that are not required with TEASER. We also evaluate on additional real-life datasets where no such alignment was performed.

We compared our approach to the state-of-the-art methods, ECTS~\cite{xing2012early}, RelClass~\cite{parrish2013classifying}, EDSC~\cite{xing2011extracting}, and ECDIRE~\cite{mori2017reliable}. On the UCR datasets, we use published numbers on accuracy and earliness of these methods to compare to TEASER's performance. As in these papers, we use $w=n_{max}/20$ as default interval length. For ECDIRE, ECTS, and RelCLASS, the respective authors also released their code, which we use to compute their performance on our additional two use-cases. We were not able to obtain runnable code of EDSC, but note that EDSC was the least accurate eTSC on the UCR data. All experiments ran on a server running LINUX with 2xIntel Xeon E5-2630v3 and 64GB RAM, using JAVA JDK x64 1.8. 

TEASER is a two-tier model using a slave and a master classifier.
As a first tier, TEASER required a TSC which produces class probabilities as output. Thus, we performed our experiments using three different time series classifiers: \emph{WEASEL}~\cite{schaefer2017weasel}, \emph{BOSS}~\cite{schafer2014boss} and 1-NN Dynamic Time Warping (DTW).
As a second tier, we have benchmarked three master classifiers, one-class SVM using \emph{LIBSVM}~\cite{chang2011libsvm}, linear regression using liblinear~\cite{fan2008liblinear}, and an SVM using an RBF kernel~\cite{chang2011libsvm}.

For each experiment, we report the evaluation metrics accuracy, earliness, their harmonic mean $HM$, and Pareto optimality. The Pareto optimality criterion counts a method as better than a competitor whenever it obtains better results in at least one metric without being worse in any other metrics. All performance metrics were computed using only results on the test split. To support reproducibility, we provide the TEASER source code and the raw measurement sheets~\cite{TEASERWebPage}.

\subsection{Choice of Slave and Master classifiers}

\begin{figure}[th]
	\includegraphics[width=1\columnwidth]{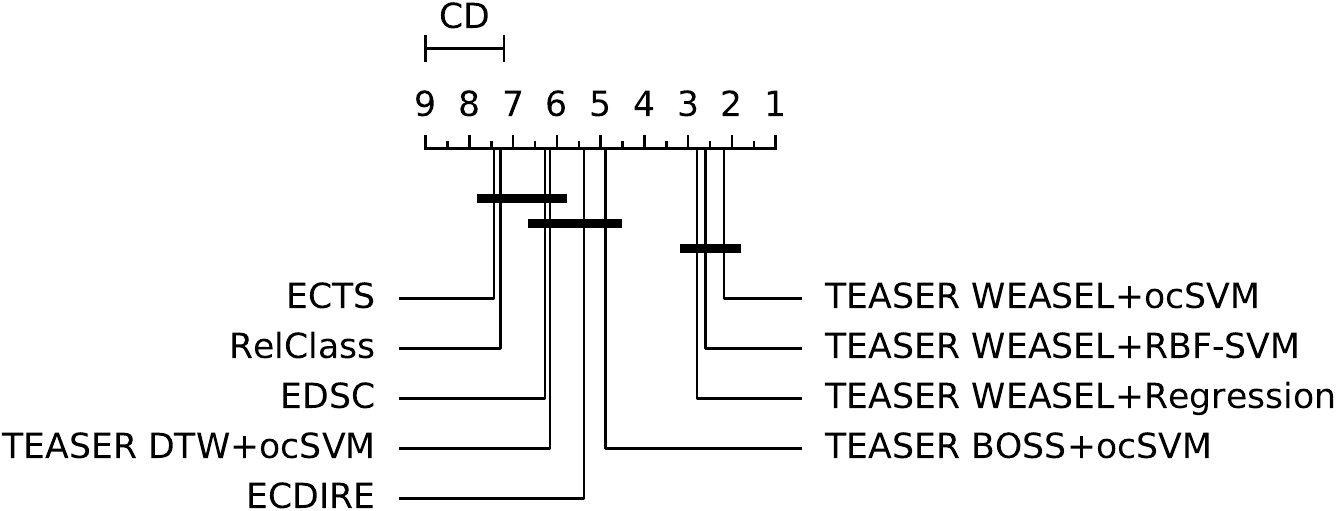}
	\caption{Average Harmonic Mean (HM) over earliness and accuracy for all 45 TS
		datasets (lower rank is better).\label{fig:hm-slaves-masters}}
\end{figure}

In our first experiment we tested the influence of different slave and master classifiers.
We compared the three different slave TS classifiers: DTW, BOSS, WEASEL. As a master classifier we have used one-class SVM (ocSVM), SVM with a RBF kernel (RBF-SVM) and linear regression (Regression). We performed these experiments using default hyper-parameters to ease comparisons. We compare performances in terms of HM to the other competitors ECTS, RelClass, EDSC and ECDIRE.

We first fixed the master classifier to oc-SVM and compared all three different slave classifiers (DTW+ocSVM, BOSS+ocSVM, WEASEL+ocSVM) in Figure~\ref{fig:hm-slaves-masters}. Out of these, TEASER using WEASEL (WEASEL+ocSVM) has the best (lowest) rank. Next, we fixed the slave classifier to WEASEL and compared the three different master classifiers (ocSVM, RBF-SVM, Regression). Again, TEASER using ocSVM performed best. The most significant improvement over the state of the art was archived by TEASER+WEASEL+ocSVM, which justifies our design decision to model early classification as a one-class classification problem.

Based on these results we use \emph{WEASEL}~\cite{schaefer2017weasel} as a slave classifier and ocSVM for all remaining experiments and refer to it as \emph{TEASER}. A nice aspect of WEASEL is that it is comparably fast, highly accurate, and works with variable length time series. As a hyper-parameter we learn the best word length between $4$ and $6$ for WEASEL on each dataset using 10-fold cross-validation on the train data. 
ocSVM parameters for the remaining experiments were determined as follows: \emph{nu-value} was fixed to $0.05$, i.e. $5\%$ of the samples may be dismissed, \emph{kernel} was fixed to \emph{RBF} and the optimal \emph{gamma} value was obtained by grid-search within $\{1 \ldots 100\}$ on the train dataset.

\subsection{Performance on the UCR Datasets}

\begin{figure*}[th]
\subfloat[Average ranks over \emph{earliness} for early TS classifiers.]{
	\includegraphics[width=0.9\columnwidth]{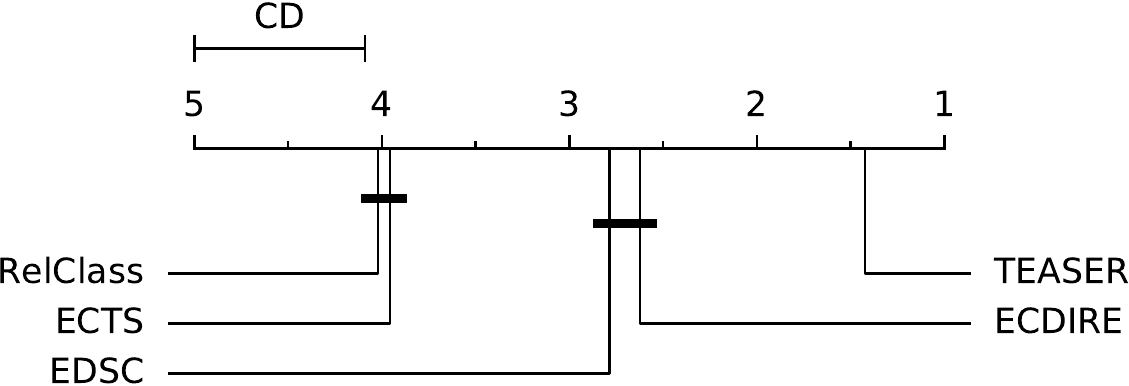}
} 
\subfloat[Average ranks over \emph{accuracy} for early TS classifiers.]{
	\includegraphics[width=0.9\columnwidth]{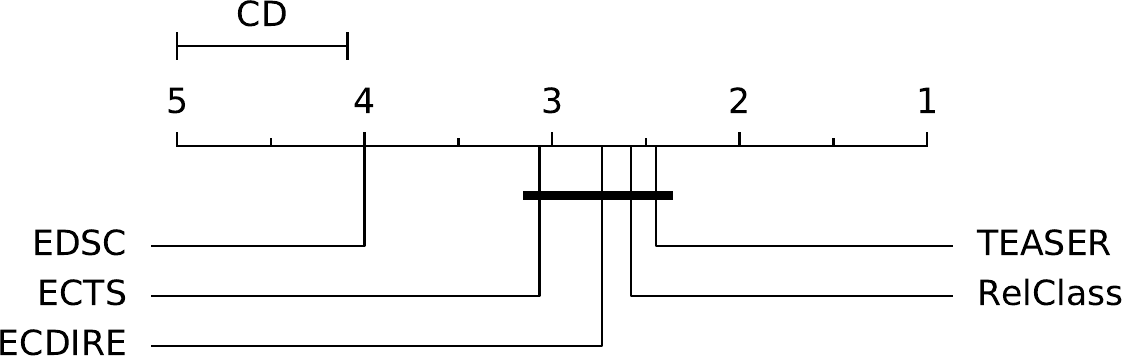}
} 
\caption{Average ranks over earliness (left) and accuracy (right) for 45 TS
datasets (lower rank is better).\label{fig:Average-ranks-over}}
\end{figure*}

\begin{figure*}[th]
	\includegraphics[width=1.8\columnwidth]{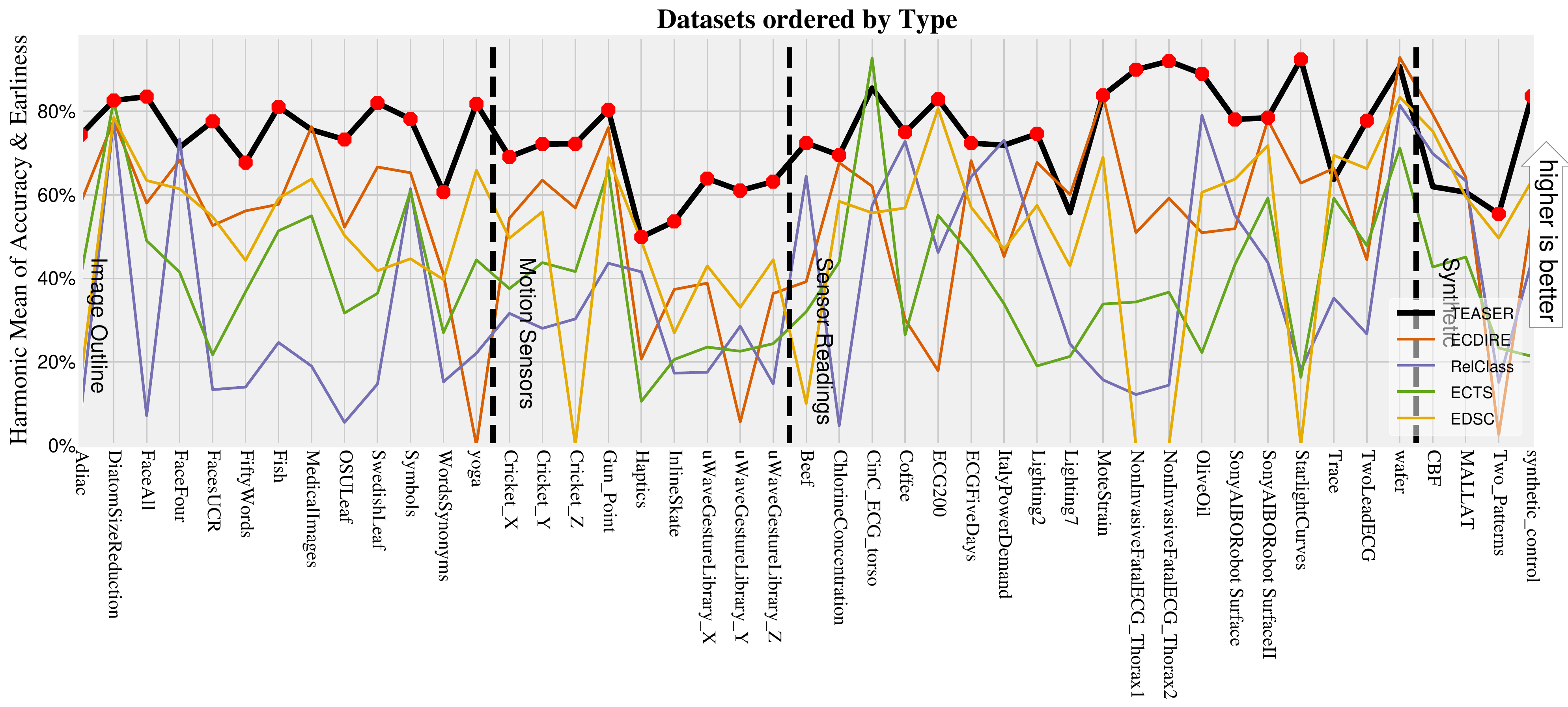}
	\caption{Harmonic mean (HM) for TEASER vs. the four eTSC classifiers (ECTS, EDSC, RelClass and ECDIRE). Red dots indicate where TEASER has a higher HM than the other classifiers. In total there are $36$ wins for TEASER.\label{fig:Classification-accuracies-for}}
\end{figure*}

eTSC is about predicting accurately and earlier. Figure~\ref{fig:Average-ranks-over} shows two critical difference diagrams (as introduced in~\cite{demvsar2006statistical}) for earliness and accuracy  over the average ranks of the different eTSC methods. The best classifiers are shown to the right of the diagram and have the lowest (best) average ranks. The group of classifiers that are not significantly different in their rankings are connected by a bar. The critical difference (CD) length represents statistically significant differences using a Wilcoxon signed-rank test. With a rank of $1.44$ (earliness) and $2.38$ (accuracy) TEASER is significantly earlier than all other methods and overall is among the most accurate approaches. 
On our webpage we published all raw measurements~\cite{TEASERWebPage} for each of the $45$ datasets.
TEASER is the most accurate on $22$ datasets, followed by ECDIRE and RelClass being best in $12$ and $10$ sets, respectively. TEASER also has the highest average accuracy of $75\%$, followed by RelClass ($74\%$), ECDIRE ($72.6\%$) and ECTS ($71\%$). EDSC is clearly inferior to all other methods in terms of accuracy with $62\%$. TEASER provides the earliest predictions in $32$ cases, followed by ECDIRE with $7$ cases and the remaining competitors with $2$ cases each. On average, TEASER takes its decision after seeing $23\%$ of the test time series, whereas the second and third earliest methods, i.e., EDCS and ECDIRE, have to wait for $49\%$ and $50\%$, respectively. It is also noteworthy that the second most accurate method RelClass provides the overall latest predictions with $71\%$. 

Note that all competitors have been designed for highest possible accuracy, whereas TEASER was optimized for the harmonic mean of earliness and accuracy (Recall that TEASER nevertheless also is the most accurate eTSC method on average). It is thus not surprising that TEASER beats all competitors in terms of $HM$ in $36$ of the $45$ cases. Figure~\ref{fig:Classification-accuracies-for} visualizes the $HM$ value achieved by TEASER (black line) vs. the four other eTSC methods. This graphic sorts the datasets according to a predefined grouping of the benchmark data into four types, namely synthetic, motion sensors, sensor readings and image outlines. TEASER has the best average $HM$ value in all four of these groups; only in the group composed of synthetic datasets EDSC comes close with a difference of just $3$ percentage points (pp). In all other groups TEASER improves the $HM$ by at least $20$ pp when compared to its best performing competitor.
\begin{table}
	\begin{centering}
		\begin{tabular}{c|cccc}
			\toprule 			
			& ECDIRE & RelClass & ECTS & EDSC\tabularnewline
			\hline 
			TEASER & 19/25/1 & 22/23/0 & 26/18/1 & 30/15/0\tabularnewline
			\bottomrule 			
		\end{tabular}
		\par\end{centering}
	\caption{Summary of domination counts (wins/ties/losses) using earliness and accuracy (Pareto Optimality):
		\label{tab:pareto}}	
\end{table}
In some of the UCR datasets classifiers excel in one metric (accuracy or earliness) but are beaten in another. To determine cases were a method is clearly better than a given competitor, we also computed the number of sets where a method is Pareto optimal over this competitor. Results are shown in Table~\ref{tab:pareto}. TEASER is dominated in only two cases by another method, whereas it dominates in $19$ to $30$ out of the $45$ cases


In the context of eTSC the most runtime critical aspect is the prediction phase, in which we wish to be able to provide an answer as soon as possible, before new data points arrives. As all competitors were implemented using different languages, it would not be entirely fair to compare wall-clock-times of implementations. Thus, we count the number of master predictions that are needed on average for TEASER to accept a master's prediction . TEASER requires $3.6$ predictions on average (median $3.0$) to accept the prediction after seeing $23\%$ of the TS on average. Thus, regardless of the used master classifier, a roughly $360\%$ faster infrastructure would be needed on average for TEASER, in comparison to making a single prediction at a fixed threshold (like  the ECTS framework with earliness of $70\%$). 

\subsection{Impact of Interval Length}\label{subsec:varying-interval-size}

\begin{figure*}[t]
\subfloat[Boxplot for \emph{earliness} for varying parameter w over all 45 datasets.]{\includegraphics[width=1\columnwidth]{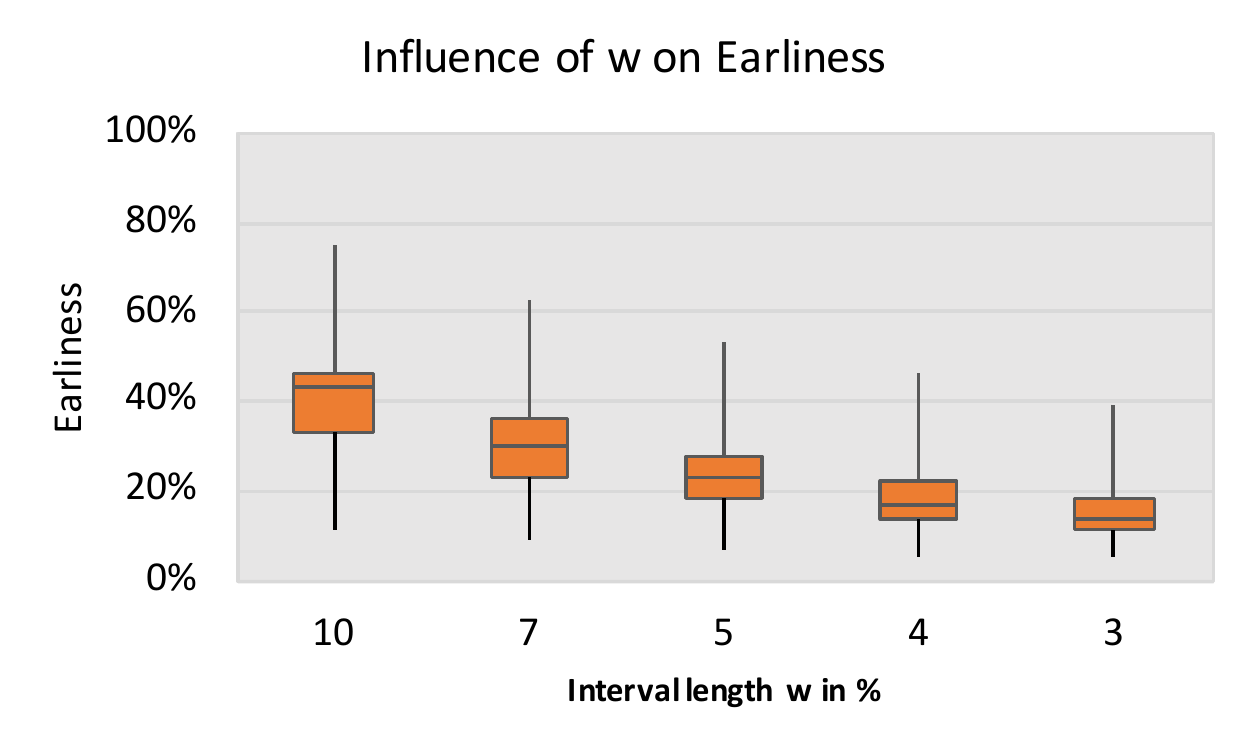}

}~~~~~~\subfloat[Boxplot for \emph{accuracy} for varying parameter w over all 45 datasets.]{\includegraphics[width=1\columnwidth]{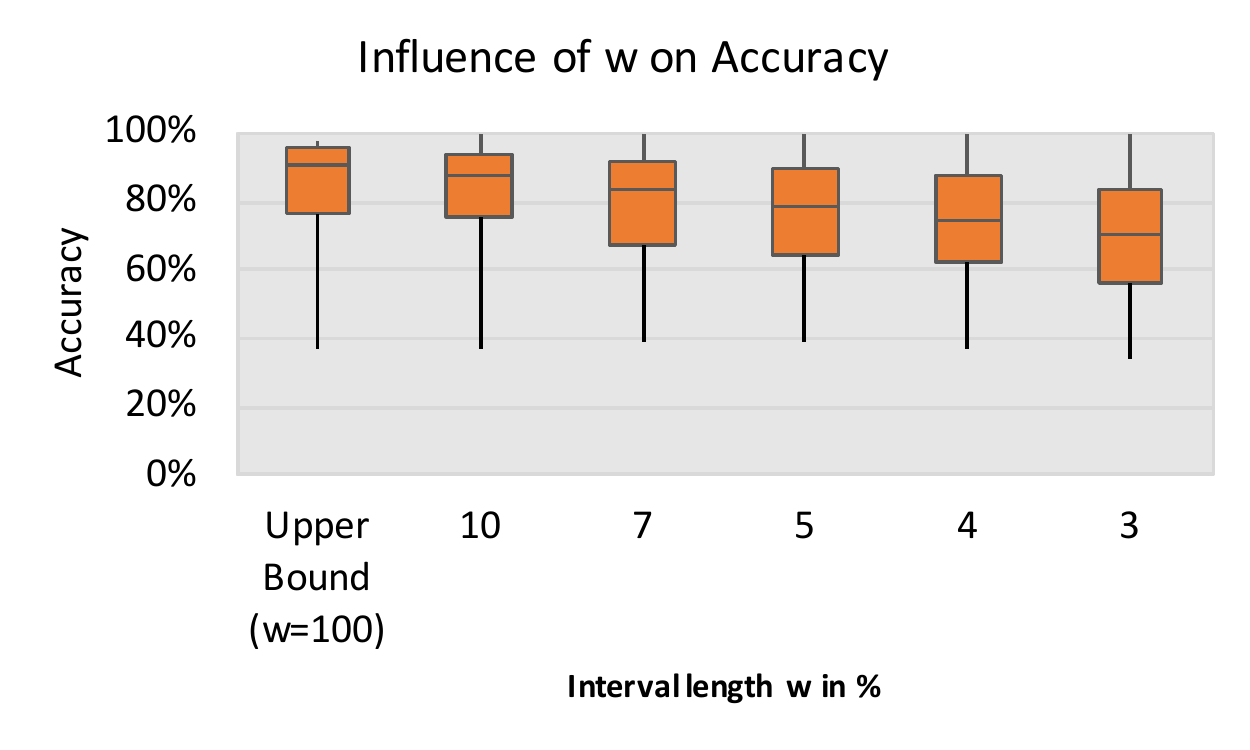}

} \caption{Average earliness (left; lower is better) and accuracy (right; higher is better) for TEASER on the 45 TS datasets.\label{fig:Average-ranks-over-1}}
\end{figure*}

To make results comparable to that of previous publications, all experiments described so far used a fixed value for the interval length $w$ derived from breaking the time series into $S=20$ intervals. Figure~\ref{fig:Average-ranks-over-1} shows boxplot diagrams for earliness (left) and accuracy (right) when varying the value of $w$ so that predictions are made after seeing multiples from $10\%$ of a dataset down to multiples of $3\%$. Thus, in the latter case TEASER outputs a prediction after seeing $3\%$, $6\%$, $9\%$, etc. of the entire time series. Interestingly, accuracy decreases whereas earliness improves with decreasing $w$, meaning that TEASER tends to make earlier predictions, thus seeing less data, with shorter interval length. Thus, changing $w$ influences the trade-off between earliness and accuracy: If early (accurate) predictions are needed, one should choose a low (high) $w$ value. 
We further plot the upper bound of TEASER, that is the accuracy at $w=100$, equal to always using the full TS to do the classification. The difference between $w=100$ and $w=10$ is surprisingly small with 5pp difference. Overall, TEASER gets to $95\%$ of the optimum using on average $40\%$ of the time series.

\subsection{Three Real-Life Datasets}

\begin{table}[t]
	\begin{centering}
		\begin{tabular}{ccccccccc}
			\toprule 
			& \multicolumn{2}{c}{Samples N} & \multicolumn{3}{c}{TS length n} & Classes \tabularnewline
			\midrule 
			& Train & Test & Min & Max & Avg & Total\tabularnewline
			\midrule
			\midrule 
			ACS-F1 & 537 & 537 & 101 & 1344 & 325 & 11\tabularnewline
			\midrule 
			PLAID & 100 & 100 & 1460 & 1460 & 1460 & 10\tabularnewline
			\midrule 
			Walking Motions & 40 & 228 & 277 & 616 & 448 & 2\tabularnewline
			\bottomrule
		\end{tabular}
		\par\end{centering}
	\caption{Use-cases ACS-F1, PLAID and Walking Motions.\label{tab:Characteristics-of-the}}
\end{table}

\begin{table*}[t]
	\begin{centering}	
		\begin{tabular}{c|ccc|ccc|ccc|ccc}
			\toprule 
			& \multicolumn{2}{c}{ECDIRE} &  & \multicolumn{2}{c}{ECTS} &  & \multicolumn{2}{c}{RelClass} &  & \multicolumn{2}{c}{TEASER} & \tabularnewline
			\midrule 
			& Acc. & Earl. & HM & Acc. & Earl. & HM & Acc. & Earl. & HM & Acc. & Earl. & HM\tabularnewline
			\midrule
			\midrule 
			ACS-F1 & 73.0\% & 44.4\% & 63.2\% & 55.0\% & 78.5\% & 31.0\% & 54.0\% & 59.0\% & 46.6\% & \textbf{83.0\%} & \textbf{19.0\%} & \textbf{82.0\%}\tabularnewline
			\midrule 
			PLAID & 63.0\% & \textbf{21.0\%} & \textbf{70.1\%} & 57.7\% & 47.9\% & 54.7\% & 58.7\% & 58.5\% & 48.6\% & \textbf{91.6\%} & 64.0\% & 51.7\%\tabularnewline
			\midrule 
			Walking Motions & 50.0\% & 68.4\% & 38.7\% & 76.3\% & 83.7\% & 26.9\% & 66.7\% & 64.1\% & 46.7\% & \textbf{93.0\%} & \textbf{34.0\%} & \textbf{77.2\%}\tabularnewline
			\bottomrule
		\end{tabular}
	\caption{Accuracy and harmonic mean (HM), where higher is better, and earliness, where lower is better, on three real world use cases. TEASER has the highest accuracy on all datasets, and the best earliness on all but the PLAID dataset.\label{tab:Accuracy-(higher-is}}
	\end{centering}
\end{table*}

The UCR datasets used so far all have been preprocessed to make their analysis easier and, in particular, to achieve roughly the same offsets for the most characteristic patterns. This setting is very favorable for those methods that expect equal offsets, which is true for all eTSC methods discussed here except TEASER; it is even more reassuring that even under such non-favorable settings TEASER generally outperforms its competitors. In the following we describe an experiment performed on three additional datasets, namely  ACS-F1~\cite{gisler2013appliance}, PLAID~\cite{gao2014plaid}, and CMU~\cite{CMU}. As can be seen from Table~\ref{tab:Characteristics-of-the}, these datasets have interesting characteristics which are quite distinct from those of the UCR data, as all UCR datasets have a fixed length and were preprocessed for approximate alignment. The former two use-cases were generated in the context of appliance load monitoring and capture the power consumption of common household appliances over time, whereas the latter records the z-axis accelerometer values of either the right or the left toe of four persons while walking to discriminate normal from abnormal walking styles.

ACS-F1 monitored about $100$ home appliances divided into $10$ appliance types (mobile phones, coffee machines, personal computers, fridges and freezers, Hi-Fi systems, lamps, laptops, microwave oven, printers, and televisions) over two sessions of one hour each. The time series are very long and have no defined start points. No preprocessing was applied. We expect all eTSC methods to require only a fraction of the overall TS, and we expect TEASER to outperform other methods in terms of earliness.

PLAID monitored $537$ home appliances divided into $11$ types (air conditioner, compact fluorescent, lamp, fridge, hairdryer, laptop, microwave, washing machine, bulb, vacuum, fan, and heater). For each device, there are two concatenated time series, where the first was taken at start-up of the device and the second during steady-state operation. The resulting TS were preprocessed to create approximately aligned patterns of equal length and scale. We expect eTSC methods to require a larger fraction of the data and the advantage of TEASER being less pronounced due to the alignment. 

CMU recorded time series taken from four walking persons, with some short walks that last only three seconds and some longer walks that last up to $52$ seconds. Each walk is composed of multiple gait cycles. The difficulties in this dataset result from variable length gait cycles, gait styles and paces due to different subjects performing different activities including stops and turns. No preprocessing was applied. Here, we expect TEASER to strongly outperform the other eTSC methods due to the higher heterogeneity of the measurements and the lack of defined start times.

\begin{figure}[th]
	\includegraphics[width=0.9\columnwidth]{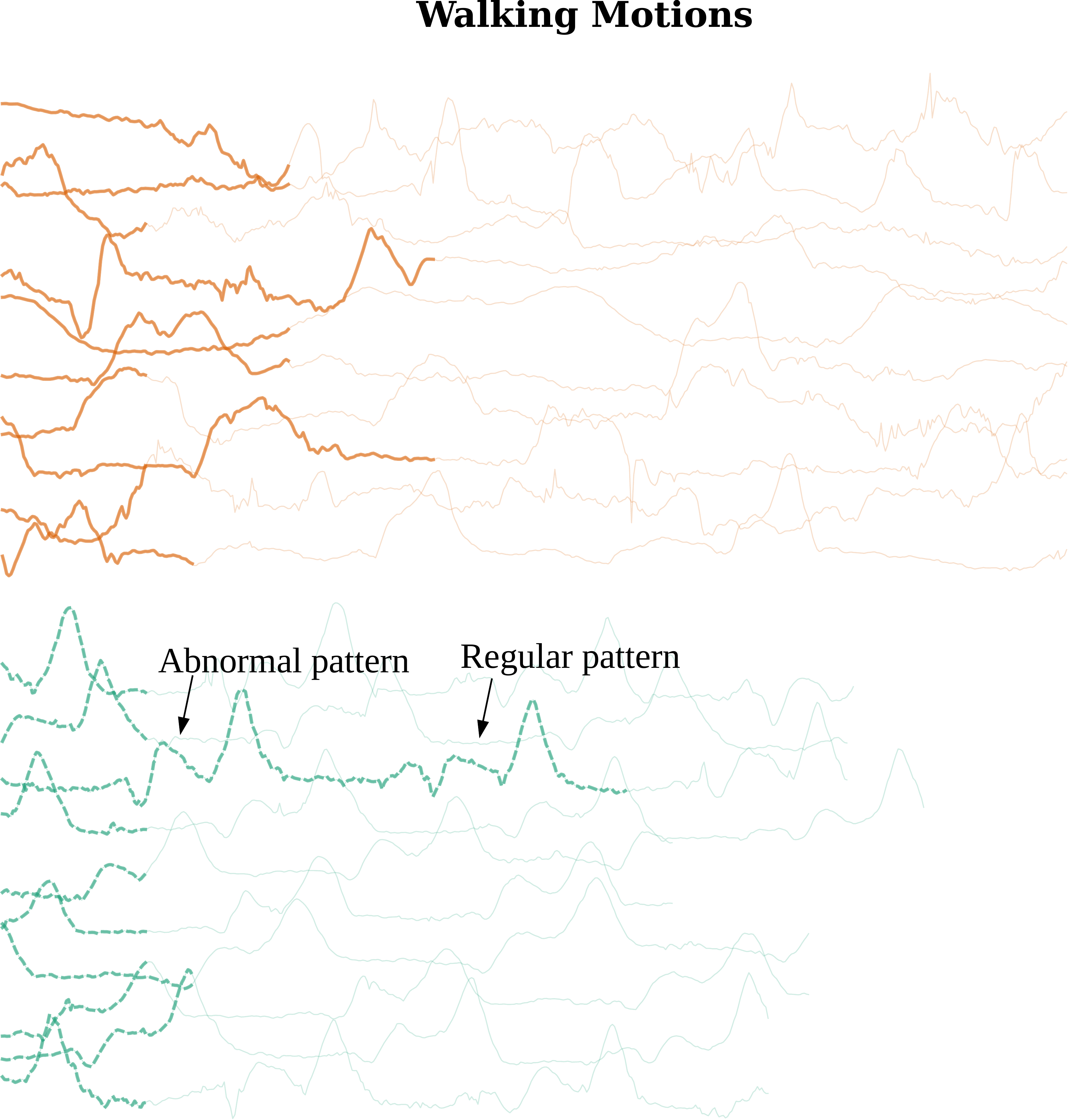}
	
	\caption{Earliness of predictions on the walking motion dataset. Orange (top): abnormal
		walking motions. Green (bottom, dashed): Normal walking motions. In bold color: the
		fraction of the TS needed for classification. In light color: the
		whole series.\label{fig:Earlinesson-the-plaid-1}}
\end{figure}

We fixed $w$ to 5\% of the \emph{maximal} time series length of the dataset for each experiment. Table~\ref{tab:Accuracy-(higher-is} shows results of all methods on these three datasets. TEASER requires $19\%$ (ACS-F1) and $64\%$ (PLAID) of the length of the sessions to make reliable predictions with accuracies of $83\%$ and $91.6\%$, respectively. As expected, a smaller fraction of the TS is necessary for ACS-F1 than for PLAID. All competitors are considerably less accurate than TEASER with a difference of $10$ to $20$ percentage points (pp) on ACS-F1 and $29$ to $34$ pp on PLAID. In terms of earliness TEASER is the earliest method on the ACS-F1 dataset but the slowest on the PLAID dataset; although its accuracy on this dataset is far better than that of the other methods, it is only third best in terms of $HM$ value. 
As  ECDIRE has an earliness of $21\%$ for the PLAID dataset, we performed an additional experiment where we forced TEASER to always output a prediction after seeing at most $20\%$ of the data, which is roughly equal to the earliness of ECDIRE. In this case TEASER achieves an accuracy of $78.2\%$, which is still higher than that of all competitors. Recall that TEASER and its competitors have different optimization goals: HM vs accuracy. Still, if we set the earliness of TEASER to that of its earliest competitor, TEASER obtains a higher accuracy.

The advantages of TEASER become even more visible on the difficult CMU dataset. Here, TEASER is $15$ to $40$ pp more accurate while requiring $35$ to $54$ pp less data points than its competitors. The reasons become visible when inspecting some examples of this dataset (see Figure~\ref{fig:Earlinesson-the-plaid-1}). A normal walking motion consists of roughly three repeated similar patterns. TEASER is able to detect normal walking motions after seeing $34\%$  of the walking patterns on average, which is mostly equal to one out of the three gait cycles. 
Abnormal walking motions take much longer to classify due to the absence of a gait cycle. Also, one of the normal walking motions (third from top) requires longer inspection time of two gait cycles, as the first gait cycle seems to start with an abnormal spike. 

\section{Related Work}

The techniques used for \emph{time series classification} (TSC) can be broadly categorized
into two classes: \emph{whole} \emph{series-based }methods and \emph{feature-based
methods~\cite{0001KL12}}. Whole series-based methods make use of
a point-wise comparison of entire TS like 1-NN Dynamic Time Warping (DTW)~\cite{rakthanmanon2012searching}.
In contrast, feature-based classifiers rely on comparing features
generated from substructures of TS. Approaches
can be grouped as either using shapelets or bag-of-patterns (BOP).
Shapelets are defined as TS subsequences that are maximally representative
of a class~\cite{YeK09,grabocka2014learning}. 
The (BOP) model~\cite{schaefer2017weasel,schafer2014boss,0001KL12, SchaferH12}
breaks up a TS into a bag of substructures, represents these substructures
as discrete features, and finally builds a histogram of feature counts
as basis for classification. 
The recent Word ExtrAction for time SEries cLassification (WEASEL)~\cite{schaefer2017weasel}
also conceptually builds on the bag-of-patterns (BOP) approach and
is one of the fastest and most accurate classifiers.  
In~\cite{wang2017time} deep learning networks are applied to TSC. Their best performing full convolutional network (FCN) performs not significantly different from state of the art. ~\cite{fawaz2018deep} presents an overview of deep learning approaches.

\emph{Early classification of time series} (eTSC)~\cite{santos2016literature} is important when data becomes available over time and decisions need to be taken as early as possible. It addresses two conflicting goals: maximizing accuracy typically reduces earliness and  vise-versa. \emph{Early Classification on Time Series (ECTS)}~\cite{xing2012early} is one of the first papers to introduce the problem. The authors adopt a 1-nearest neighbor (1-NN) approach
and introduce the concept of minimum prediction length (MPL) in combination
with clustering. Time series with the same 1-NN are clustered. The
optimal prefix length for each cluster is obtained by analyzing the
stability of the 1-NN decision for increasing time stamps. Only those
clusters with stable and accurate offsets are kept. To give a prediction
for an unlabeled TS, the 1-NN is searched among clusters. \emph{Reliable
Early Classification (RelClass)}~\cite{parrish2013classifying} presents
a method based on quadratic discriminant analysis (QDA). A reliability score
is defined as the probability that the predicted class for the truncated
and the whole time series will be the same. At each time stamp, RelClass
then checks if the reliability is higher than a user-defined threshold.\emph{
Early Classification of Time Series based on Discriminating Classes
Over Time (ECDIRE)}~\cite{mori2017reliable} trains classifiers at
certain time stamps, i.e. at percentages of the full time series length.
It learns a \emph{safe} time stamp as the fraction of the time series
which states that a prediction is safe.
Furthermore, a reliability threshold is learned using the difference between the
two highest class probabilities. Only predictions passing this threshold
after the \emph{safe} time stamp are chosen. 
The idea of \emph{EDSC~}\cite{xing2011extracting} is to learn Shapelets that appear early in the time series, and that discriminate between classes as early as possible. 
\cite{mori2017early} approaches early classification as an optimization problem. The authors combine a set of probabilistic classifiers with a stopping rule that is optimized using a cost function on earliness and accuracy. 
Their best performing model \emph{SR1-CF1} is significantly earlier than the state
of the art but their accuracy falls behind. However, the code has a logical design flaw, which renders the results hard to compare to, but apparently results in good scores on the UCR datasets. Their algorithm uses z-normalized time series, which are then truncated to build a training set. Thereby, a truncated subsequence makes use of information about values that will only be available in the future for normalization. I.e., their absolute values are a result of z-normalization with data that has not arrived yet. In TEASER the truncated train series are z-normalized first, thus removing any bias from values that have not been seen. We decided to omit SR1-SF1 from our evaluation due to this normalization issue.
A problem related to eTSC is the classification of streaming time series~\cite{Nguyen2015, gaber2005mining}. In these works, the task is to assign class labels to time windows of a potentially infinite data stream, and is similar to event detection in streams~\cite{aggarwal2012event}. The data enclosed in a time window is considered to be an instance for a classifier. Due to the windowing, multiple class labels can be assigned to a data stream. 
In contrast, eTSC aims at assigning a label to an entire TS as soon as possible. 

\section{Conclusion}

We presented TEASER, a novel method for early classification of time series. TEASER\textquoteright s decision for the safety (accuracy) of a prediction is treated as a classification problem, in which master classifiers continuously analyze the output of probabilistic slave classifiers to decide if their predictions should be trusted or not. By means of this technique, TEASER adapts to the characteristics of classes irrespectively of the moments at which they occur in a TS. In an extensive experimental evaluation using altogether $48$ datasets, TEASER outperforms all other methods, often by a large margin and often both in terms of earliness and accuracy.

\balance

\bibliographystyle{abbrv}
\bibliography{early}


\end{document}